\documentclass[sn-mathphys,Numbered]{sn-jnl}% Math and Physical Sciences Reference Style
\usepackage{graphicx}%
\usepackage{multirow}%
\usepackage{amsmath,amssymb,amsfonts}%
\usepackage{amsthm}%
\usepackage{mathrsfs}%
\usepackage[title]{appendix}%
\usepackage{xcolor}%
\usepackage{textcomp}%
\usepackage{manyfoot}%
\usepackage{booktabs}%
\usepackage{algorithm}%
\usepackage{algorithmicx}%
\usepackage{algpseudocode}%
\usepackage{listings}%

\usepackage{natbib}
\usepackage{url}

\usepackage{subcaption}
\usepackage{todonotes}
\usepackage{svg}
\theoremstyle{thmstyleone}%
\theoremstyle{thmstyletwo}%

\theoremstyle{thmstylethree}%

\begin{document}

\title[Article Title]{Combining Embeddings and Domain Knowledge for Job Posting Duplicate Detection}
%\title[Article Title]{A Job Posting Duplicate Detection Application combining Embeddings and Domain Knowledge}
%Original: Titel ist so lang in dem Templte, wir sollten vllt mit einem prägnanteren aufwarten 
%\title[Article Title]{Combining String Similarity with Embeddings and Domain Knowledge for Duplicate Detection in Text Documents -\\ A Deployed Application for Job Postings Analysis}

%%=============================================================%%
%% Prefix	-> \pfx{Dr}
%% GivenName	-> \fnm{Joergen W.}
%% Particle	-> \spfx{van der} -> surname prefix
%% FamilyName	-> \sur{Ploeg}
%% Suffix	-> \sfx{IV}
%% NatureName	-> \tanm{Poet Laureate} -> Title after name
%% Degrees	-> \dgr{MSc, PhD}
%% \author*[1,2]{\pfx{Dr} \fnm{Joergen W.} \spfx{van der} \sur{Ploeg} \sfx{IV} \tanm{Poet Laureate} 
%%                 \dgr{MSc, PhD}}\email{iauthor@gmail.com}
%%=============================================================%%

\author*[1]{\fnm{Matthias} \sur{Engelbach}}\email{matthias.engelbach@iao.fraunhofer.de}
\equalcont{These authors contributed equally to this work.}

\author[2]{\fnm{Dennis} \sur{Klau}}\email{dennis.klau@iat.uni-stuttgart.de}
\equalcont{These authors contributed equally to this work.}

\author[1]{\fnm{Maximilien} \sur{Kintz}}\email{maximilien.kitz@iao.fraunhofer.de}

\author[3]{\fnm{Alexander} \sur{Ulrich}}\email{alexander.ulrich@contractor.de}

%\affil*[1]{{\orgdiv{Institute for Industrial Engineering IAO}}}
\affil*[1]{\orgdiv{Institute for Industrial Engineering IAO}, \orgname{Fraunhofer}, \orgaddress{\street{Nobelstrasse 12}, \city{Stuttgart}, \postcode{70569}, \country{Germany}}}

%\affil[2]{\censor{\orgdiv{Institute of Human Factors and Technology Management IAT}}}
\affil[2]{\orgdiv{Institute of Human Factors and Technology Management IAT}, \orgname{University of Stuttgart}, \orgaddress{\street{Nobelstrasse 12}, \city{Stuttgart}, \postcode{70569}, \country{Germany}}}

%\affil[3]{\censor{\orgname{Contractor Consulting GmbH}, \orgaddress{\city{Munich}, \postcode{80797}, \country{Germany}}}}
\affil[3]{\orgname{Contractor Consulting GmbH}, \orgaddress{\street{Adams-Lehmann-Str. 56}, \city{Munich}, \postcode{80797}, \country{Germany}}}

%%==================================%%
%% sample for unstructured abstract %%
%%==================================%%

\abstract{
%\todo[inline]{TODOs: Alle Referenzen auf uns und Contractor im Text zensieren; Anschriften sauber zensieren; Titel ggf. überarbeiten}
%Job descriptions are posted on many online channels, including company websites, job boards or social media platforms such as \name{LinkedIn}. Often, these descriptions are published with different text variations for the same job, due to the requirements of each platform or to target different audiences. However, for the purpose of automating recruitment and assisting people working with these job descriptions, it is helpful to aggregate job postings across platforms and thus detect duplicate descriptions that refer to the same job. In this work, we propose an approach for detecting duplicates in job descriptions. We show that combining overlap-based character similarity with text embedding and keyword matching methods lead to convincing results. In particular, we show that although no approach individually achieves satisfying performance, a combination of string comparison, deep textual embeddings, and the use of curated weighted lookup lists for specific skills leads to a significant boost in overall performance. A tool based on our approach is being used in production and feedback from real-life use confirms our evaluation.
Job descriptions are posted on many online channels, including company websites, job boards or social media platforms. These descriptions are usually published with varying text for the same job, due to the requirements of each platform or to target different audiences. However, for the purpose of automated recruitment and assistance of people working with these texts, it is helpful to aggregate job postings across platforms and thus detect duplicate descriptions that refer to the same job. In this work, we propose an approach for detecting duplicates in job descriptions. We show that combining overlap-based character similarity with text embedding and keyword matching methods lead to convincing results. In particular, we show that although no approach individually achieves satisfying performance, a combination of string comparison, deep textual embeddings, and the use of curated weighted lookup lists for specific skills leads to a significant boost in overall performance. A tool based on our approach is being used in production and feedback from real-life use confirms our evaluation.
}


%%\pacs[JEL Classification]{D8, H51}

%%\pacs[MSC Classification]{35A01, 65L10, 65L12, 65L20, 65L70}

\maketitle

\section{Introduction}
\label{sec:introduction}

With increasing digitization of business processes, the possibility to assist people working in the area of human resources by leveraging means of automated recruiting has become an important topic~\cite{donovan2017automated,leong2018technology}. Technologies like artificial intelligence (AI) and robot process automation (RPA) can be applied to a wide range of tasks in a typical recruiting process, such as automated screening of job portals (e.g. LinkedIn) for retrieving job offerings published by hiring companies or personal job profiles and CVs~\cite{kong2021ai}.

However, even with advanced methods and tools for collecting such data, further processing can be challenging. Especially when merging data stemming from multiple sources, curating and cleaning the vast amount of records captured regularly is crucial~\cite[p.~6]{draisbach2022efficient}. 
For instance, a database might contain current job postings targeting recruitment of freelancers for IT projects that have been published on several job platforms in parallel and which in fact relate to one single job reference. 
As of now, while many other steps in the recruitment process can already be automated quite successfully, the detection of such duplicates within unstructured textual data is not an easy task to solve~\cite{ramya2016feature}. 
This is partially due to the fact that platform and company specific constraints often result in similar, but not identical postings for both: different job advertisements being part of the same project, as well as job postings coming from the same company, but relating to different projects~\cite{burk2017apollo}.

For those reasons simple automated duplicate checks usually lead to a high rate of false positives, which is why this task is often still solved in the form of time consuming manual work. To address this problem and to provide a useful programmatic solution, we investigated different methods of similarity detection tailored to the specific goal of finding duplicates among the textual contents of job postings.

Our key findings are that several approaches deliver good results, but only a combination of text similarity matching with domain knowledge in the form of a weighted skills averages and text embedding gives a very high score with few false positives.

An application based on our duplicate detection approach was developed in a economically viable way and has been successfully deployed as a live system for the past months.

\section{Industrial Use Case and Requirements}
\label{sec:requirements}

Recruiters deal with the matching of candidates with job positions, for example finding a suitable IT freelancers for specific software development projects. 
One task the recruiters need to perform consists in the curation of a job listings database. 
Companies looking for IT freelancers often publish open positions on many different web sites, in slightly different variations. 
These open positions need to be aggregated and duplicates need to be identified. 
This time-consuming process occurs mostly manually or with some degree of automation that however is error prone and requires manual oversight.

To give an idea of the difficulty of the task, some example borderline cases can be named:
\begin{itemize}
    \item the same job posting may be published on two different sites with slightly different wording in the title and text description,
    \item the same company may be looking for two people for the same Java development project, one for front-end and one for back-end, resulting in very similar postings for different jobs or
    \item the same employer may be looking for two Java developers in a project, one working on site in the company offices, one being a remote working position.
\end{itemize}
These examples show that detecting duplicate job listings is hard to achieve in a reliable way by simply matching basic data such as job title description and company name, or only analyzing the required skills. A sophisticated combination of several different criteria, however, can greatly increase the detection performance.

As the project was conducted for a real-world use-case in cooperation with Contractor Consulting GmbH\footnote{\url{https://www.contractor.de}}, the goal was to have a tool ready for production use. Because of this, some specific requirements applied that may would not have been relevant in the setting of a larger research project:
\begin{itemize}
    \item Both, the development and operation costs had to be limited in order to get a viable business case. The development and usage of the resulting tool must be cheaper than the labor cost of manually detecting duplicates, and the additional costs generated by false detection (false positives/negatives). The tool will be used in a semi-automated fashion, where listings marked as duplicate, i.e. true and false positives, are reviewed by the recruiting experts; while non-duplicates are processed automatically and no further review or action is triggered. Therefore, in the context of misclassifications, false positives pose less of an issue to the users than false negatives due to the human review and feedback loop. The potential savings generated by the new tool do not justify the development and usage of large-scale solutions.
    Development effort was limited to about 15 to 20 person days.
    \item Additionally, the software shall have no licensing and hardware renting costs. This means that using large cloud AI platforms or hardware-intensive large language models was not an option.
    \item Whenever possible, easy to understand and easy to maintain solutions should be preferred, so as to ensure that the tool can be maintained by the end users.
\end{itemize}
These requirements constrained the scope of considered approaches and tools used during the project, also excluding the fine-tuning of big language models with curated data sets.
%These requirements constrained the scope of the reviewed related work and of the tools that were used during the project, also excluding any kind of fine-tuning approach based on machine learning.

\section{Related Work}
\label{sec:related}

Usage of AI solutions in recruitment processes is common and well documented, though typically the task addressed is the analysis and review of applications and resumes \cite{Chan2022}.

Some research has already addressed job descriptions. In \cite{vo2021dealing}, the authors show how removing stop words and extracting relevant features from online job description postings help in the detection of fake postings.

Works like \cite{burk2017apollo} address the specific task of duplicate detection for online recruitment using standard de-duplication techniques like Jaccard Similarity and n-grams with limited success. Furthermore, \cite{zha021} investigated different 24 methods for job duplicate detection, including usage of different tokenizers. 
Another common use case from a different, yet still technical domain is the detection of duplicates among free text bug reports, where in some cases similar techniques have been applied \cite{sureka10} \cite{tian12}
% Zusätzlich wenn man möchte  \cite{sun11} (z.T. gleicher Autor wie tian12).
However, extensive usage of domain knowledge is not applied there.  

Text comparison is a common task for which many approaches exist. However, classic algorithms such as Levenshtein distance \cite{Levenshtein66}, TF-IDF \cite{tfidf} or longest common subsequence (LCS) \cite{longestCommonSub} are not suitable for our approach, as the job listings contain many similar-looking text blocks on a word and character basis, but are often very different in terms of meaning.

\citet{shortTextWord2Vec} use Word2Vec \cite{word2vec13} embeddings to detect short text duplicates. The usage of embeddings is clearly relevant to our work, job descriptions tend to be long documents and thus require a different approach than the one proposed by the authors. 

\citet{plagiarismReview} provide a review of plagiarism detection methods, an area related but not identical to our problem domain. The findings confirm that using new machine learning techniques and a combination of approaches is most promising.

\citet{similarityMeasurement} give an overview of text similarity measurement techniques. One main finding of the study that is relevant to the present work is the confirmation that text matching is application and domain-specific, i.e. dependent on the type of texts and on the task to be accomplished after the similarity has been computed. In order to create performant solutions, different algorithms need to be considered.

Overall, our overview of related work shows that though many approaches exist for computing text document similarity, a selection of the most appropriate one is always task and domain specific and thus needs to be reviewed for each new project. The particularities of job descriptions texts and the project requirements mentioned in the previous section lead us to conduct the experiments described in the following chapters.

\section{Data Resources}
\label{sec:data}
During the project phase we had access to a database with about 45.000 collected project tenders and according textual descriptions. The typical content of a job description is well established and has already been studied. \cite{contentOfJobDescriptions} mention the main elements of such a text: a job title, required skills, preferred skills and responsibilities. In our specific case and dataset, we identified the following features as most relevant:

\begin{itemize}
    \item \textbf{Title:} The headline on top of the project proposition
    \item \textbf{Description:} The text body describing the project and job requirements
\end{itemize}
Additionally, each project in the data set is identified by a unique project ID.
The \textit{description} is the textual main body of the job posting and typically consists of the following elements: 
\begin{itemize}
    \item a short \textit{introduction} text usually describing the company that is looking for a freelancer,
    \item the \textit{tasks to be accomplished} -- often in form of a bullet-point list,
    \item \textit{requirements to be met} by the potential candidate, for example years of experience in a specific area, required hard and soft skills, etc.
    \item \textit{general working information} such as start and end date of the project, remote or on-site position, or amount of hours of work expected
    \item a short \textit{ending} text describing the application process.
\end{itemize} 

The exact elements and their order may vary from document to document. Additionally, many postings contain additional "glue" or "boilerplate" text added by the company or recruiting agency, which does not carry information for our task, but can lead to false positive detections.

Besides the textual information that we get for each job offer, the data set also contains some meta information for each sample like date of publishing, which we use to limit matching candidates to a realistic time range in which we expect duplicates of the same job to appear (e.g., a few weeks). Other features like working place, duration (for freelancers that work on specific projects this is usually already fixed in the job posting) and starting date are also stored for every job posting.
Although these structured features could also be indicators for duplicates, during our pretesting we found them to be misleading for many cases as well, since different projects and jobs might share the same place or dates and some features are missing for some candidates in the database. Even though such information might be useful for the validation of detected duplicates, in our work we focus on comparing the larger textual features mentioned before such as job description and title.

In addition to the job posting data, we had access to an (automatically created) key word list containing the $25.000$ most frequent skills (occurring at least seven times in resumes). Accompanied by the skill list we also introduced a \textit{skill black list} during the project phase with the aim to exclude misleading or wrong entries in the skill list.

A small test set was manually annotated with $50$ \textit{duplicate-pairs} identified by Contractor. This test set exhibits the a high quality and was used only used for final testing of the deployed pipeline.
After the implementation of the first version of the detection pipeline, we created a validation set by annotating a set of $176$ job posting-pairs with corresponding predictions by the pipeline. After curation, we ended up with $74$ duplicates and $73$ non-duplicates. We refer to those $73$ non-duplicates also as fake-duplicates, since they were labeled as duplicates by our algorithm and therefore exhibit at least some degree of similarity to the matching source. $29$ of the predictions could not be assigned confidently to one of the classes (duplicate, non-duplicate) even after review from the domain experts and were removed from the curated dataset. This validation set was used to fine-tune the parameters and thresholds during development of the matching algorithms.

%For the purpose of textual similarity matching, we made use of the different data resources, especially performing string and embedding comparisons for the textual data set features \emph{Description} and \emph{Title} in combination with the skill lists and method evaluations on the test sets.

\begin{figure*}[htb]
  \centering
   \includegraphics[width=\textwidth]{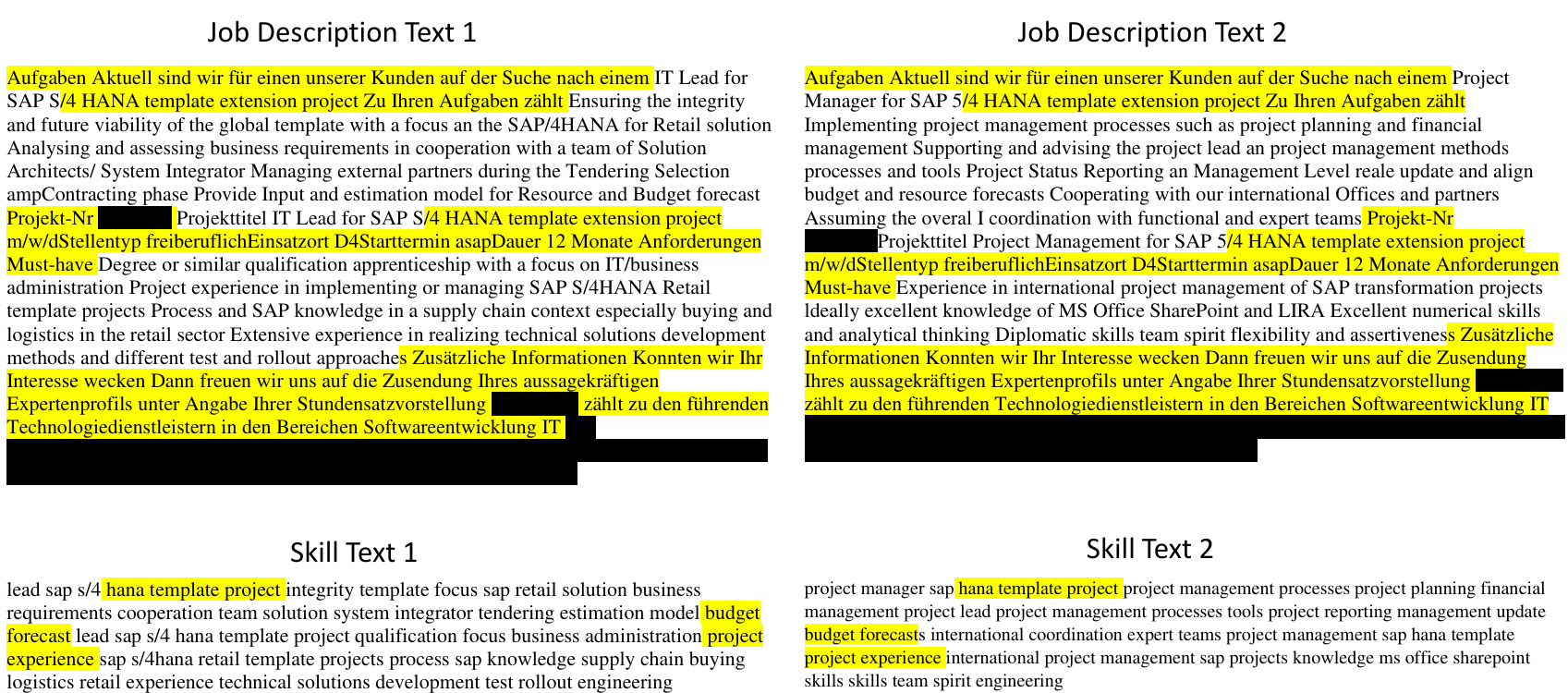}
  \caption{Example of job offering comparison for two different positions that are part of the the same project. While the original description text has greater parts of overlap in boiler plate text sections, the extracted pure skill texts differ significantly. Yellow markings indicate overlapping text parts between source and target texts. The censored parts mark confidential information.}
  \label{fig:overlap}
 \end{figure*}
 
\section{Method and Experimental Setting}
\label{sec:method}

\subsection{Data Preprocessing}
Before comparing job offerings with each other, the whole dataset (and especially the textual contents) are preprocessed and cleaned: we remove special characters such as line breaks or enumeration characters to receive a normalized version of all texts -- mainly containing letters, white spaces and numbers. We make an exception for special characters like '+' or slashes that can appear in technical skills (e.g. C++, S/4 Hana), which we keep in the text corpus. Besides, similar to the approach proposed by \citet{gunawan2018implementation}, we extract and lowercase all skill terms from the original description texts based on the skill list and skill black list mentioned in the previous section together with their weights (see Section "Weighted Keyword Matching"). This preprocessing is done for all further matching methods and score computations.

\subsection{Evaluation Metrics}
\label{metrics}
For measuring the quality of our matching algorithms, we resort to standard metrics like Precision, Recall and F1-Score, which have proven to be reliable indicators for duplicate detection tasks \cite{bilenko2003evaluation}. We compute them over all matched pairs of project publication records in our data sets. In detail, we handle a comparison result as true positive (TP) if an actual duplicate has been detected successfully, and as false negative (FN) if this is not the case. Similarly, non-duplicate pairs that have been assigned with a duplicate label are treated as false positives (FP), while true negatives (TN) are all pairs of correctly labeled non-duplicates.
As proposed by previous works \cite{SemanticThreshold}, we use score dependent threshold values (TH) for the decision of label assignment. Thus, if a matching pair sample gets assigned a score above the threshold by an algorithm, it is labeled as a predicted duplicate, otherwise it is assigned the non-duplicate prediction label.
We use the validation set containing both duplicates and fake-duplicate job pairs introduced in the previous section as a development set for score and threshold optimization and evaluate our final pipeline on the test set containing only known duplicate pairs to ensure that our approach is valid for unseen data.

\subsection{Overlap-based Similarity Matching}
\label{subsec:overlap}

We use a simple \textit{Text Overlap Score} computation, that computes the degree of matching text blocks between two job publications and serves as a baseline for all further experiments. In detail, the Python package \emph{levenshtein}\footnote{\url{https://maxbachmann.github.io/Levenshtein/levenshtein.html}} is used to get all larger text blocks (with minimum length of 15 characters) within the source text that also appear in the target candidate of the current matching pair.

We define $S$ as the source text and $T$ to be the target matching text. For each overlapping block $i$, we define $b_{i}$ as its length. If there are a total of $n$ overlapping blocks, the total length of overlapping text is:

\[ B = \sum_{i=1}^{n} b_{i} \]

\noindent Then, the $TOS$ is calculated as follows:

\[ TOS = \frac{B}{\text{len}(S)} \]

\noindent This means, we sum up the length of the overlapping string blocks (number of overlapping characters) and normalize the result by dividing through the total length of the source text to get a percentage value that indicates the degree of overlap.

During our first pretest experiments using the overlap score, we found that many of the job offering texts consist of larger blocks of boiler plate text (such as introduction of the company or common contact information) that are shared among different job offerings published by the same company or agency. This results in high $TOS$ values for different job postings. On the other hand, we identified other passages of the description text as most relevant for distinguishing between job offers, specifically all sections containing project-related skill requirements.

In Figure~\ref{fig:overlap} we illustrate this finding with an example: The start and end of the matching source text overlap in most parts with the target, while the skill list indicates that in this case indeed different job offerings are compared. 

For this reason, besides computing a simple overlap score over the whole job description text, we additionally compute a \textit{Skill Overlap Score} ($SOS$) that only considers matching blocks appearing in the concatenation of all listed skill words and terms detected in the whole description text (see bottom of Figure~\ref{fig:overlap}). These pure skill texts, representing the most specific parts of a job offering, are extracted during the preprocessing phase by using the skill list.

Note, that for each pair matched with the overlap scores we compute a forward and a backward score, since the choice of source and target texts influences the length of the resulting matching blocks. Therefore, the final $TOS$ is calculated by taking the mean of the forward and backward pass:

\[ TOS_{\text{final}} = \frac{TOS_{\text{forward}} + TOS_{\text{backward}}}{2} \]

\noindent The final Skill Overlap Score $SOS$ is computed analogously.

 \begin{figure*}[tb]
  \centering
   \includegraphics[width=\textwidth]{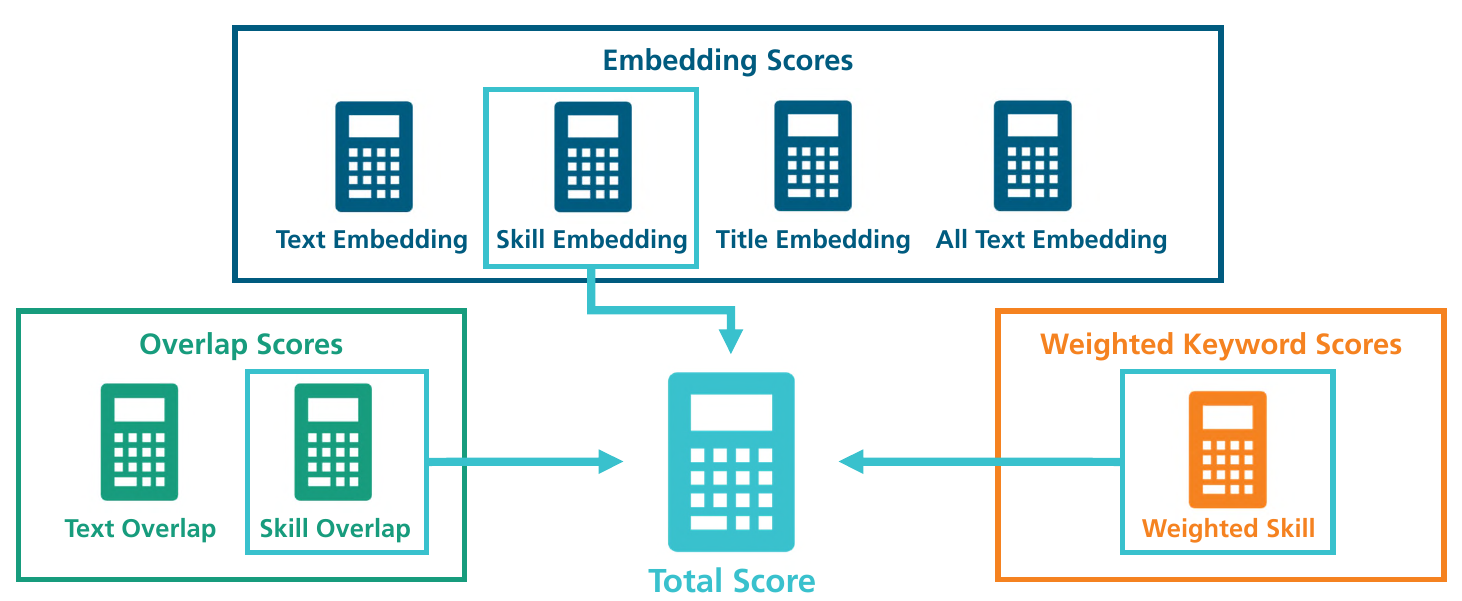}
  \caption{Overview of the different scoring approaches in their respective matching classes and the final selected component of each class for the total score. The components were selected by their performance on the validation set.}
  \label{fig:scoreoverview}
 \end{figure*}
 
\subsection{Embedding-based Similarity Matching}
\label{subsec:embedding}
In addition to the string overlap scores, we also test similarity matching based on text embeddings in order to capture semantic similarities (e.g. different wordings and phrases) not detected by the strict string matching approach. 
In detail, since our data contained both German and English job descriptions, we used a multilingual text encoding method as proposed by \citet{yang2019multilingual} and provided by the pretrained model \emph{distiluse-base-multilingual-cased-v1}\footnote{\url{https://www.sbert.net/docs/pretrained_models.html}} from the Python library \emph{sentence-transformers} \cite{DBLP:journals/corr/abs-2004-09813}.

Embeddings for all data samples are computed during preprocessing and cached for all further comparison actions. During pairwise matching of job descriptions, we compute the \textit{Embedding Score} ($ES$) with the commonly used cosine similarity \cite{singhal2001modern} between embedding vectors $S_e$ of source $S$ and $T_e$ of target candidate $T$ as follows:

\begin{equation*}
    ES = \text{cosine similarity}(S_e, T_e) = \frac{S_e \cdot T_e}{\|S_e\| \|T_e\|}
\end{equation*}

\noindent We then normalize the results by mapping (eventually appearing) negative values to 0 (indicating no significant similarity between $S$ and $T$) for ensuring comparability of scores. 

Similar to the computation of the overlap scores, we calculate embedding scores for different parts of the original job description:
\begin{itemize}
    \item \textit{Text Embedding Score ($TES$)}: Embedding over the whole job description text
    \item \textit{Skill Embedding Score ($SES$)}: Embedding over the pure skill text (concatenation of all found skill terms)
    \item \textit{Title Embedding Score ($TTES$)}: Embedding over the headline of the job description (retrieved from a separate feature in the data set)
    \item \textit{All Text Embedding Score ($AES$)}: Embedding over concatenation of all texts (title, description and skill text)
\end{itemize}

\subsection{Weighted Keyword Matching}
\label{subsec:weighted}
During the pretesting phase of the project, we found that skill terms mentioned in job descriptions are key features regarding the task of duplicate detection. On the other hand, since the list of relevant skills was created in an automated manner, it also contains wrong or misleading keywords (examples for that are "English", "experience" or "team work") that are not indicating real technical skill requirements for the job. Furthermore, very common terms like "Java", that are occurring in a lot of job descriptions, are considered as not helpful for finding duplicates. Although a skill black list as introduced in the Data Resources section was used, manual capturing of all misleading terms was not feasible within the scope of the project.

For those reasons and as proposed by other work like \cite{gunawan2018implementation}, we introduce another matching approach: a \textit{Weighted Skill Score} ($WSS$). In contrast to the previous, presented scores which use the skill list to different degrees, the $WSS$ focuses on the significance of individual skill keywords by taking into account their relevance among the whole corpus of the data set. 
For that, we calculate weights for every skill term in the skill list during preprocessing by counting their occurrences in all project descriptions of the data set to approximate a value of their significance.
The weights are re-computed on a regular basis, witch could also be done in an online manner while the system is active and new project descriptions are added to the data base. This way the system can continuously learn during its operation time. 
For a skill term $s_i$ from the skill list, we define $f(s_i)$ as its frequency in the entire project database. Then, the weight of the skill term is:

\[ w(s_i) = \frac{1}{f(s_i)} \]

\noindent During calculation of the weighted score for a matching pair, we take the inverse of every precomputed skill term's frequency because we assign more frequent skill words less relevancy for duplicate detection than specific skills that appear very seldom.

For a source text $S$ and a target candidate $T$, the \textit{Weighted Skill Score} $WSS$ is then calculated as follows:

\[ WSS = \frac{\sum_{s_i \in (S \cap T)} w(s_i)}{\sum_{s_i \in S} w(s_i)} \]

\noindent The $WSS$ consists of the sum of weights of all skill terms found in both job descriptions $S$ and $T$ divided by the sum of weights of all skills within the source text of the matching pair.

Again, we perform a forward and backward pass by swapping $S$ and $T$ and use the mean of both for the final weighted skill scores $WSS_{\text{final}}$:

\[ WSS_{\text{final}} = \frac{WSS_{\text{forward}} + WSS_{\text{backward}}}{2} \]

\subsection{Combined Matching}
\label{subsec:combined}
As a result of the experiments during a first pretesting phase, we deducted a final combined \textit{Total Score} ($TS$) that is calculated by averaging over the three (best performing) individual scores -- each coming from one of three categories mentioned before. Figure~\ref{fig:scoreoverview} visualizes the score influences contributing to our total score: we found that the skill related metrics perform best during our tests, thus the $TS$ consists of the average of Skill Overlap Score $SOS$, Skill Embedding Score $SES$ and Weighted Skill Score $WSS$. In principal, all employed scoring methods are language and location independent and have been applied to German and English job postings. For better performance of the embedding methods, it is advised to use an appropriate embedding model for the target language or a multilingual one.

%\[ TS = \frac{SOS + SES + WSS}{3} \]

\section{Results and Discussion}
\subsection{Threshold Finding}
To identify suitable thresholds for each score, we start visualizing the prediction results for the different methods on the validation set with duplicates and (similar looking) non-duplicates. Figure \ref{fig:scores} shows those predictions for every scoring method introduced in the previous section. Note that, due to the constrained scope of the project, we set the threshold values manually by human supervision to achieve a good prediction performance for each scoring method. However, we use the same values for every method group to preserve comparability. The thresholds are listed in Table~\ref{tab:results} together with the according score performance when applying these threshold values for prediction on the validation set.

\subsection{Experimental Results} 

\begin{table*}[bt]
%\vspace{-0.2cm}
\centering
\caption{Experimental results for different matching score computations listed as Precision (P), Recall (R) and F1-Score (F1) of correctly recognized project duplicates on the validation set. TH indicates the threshold chosen for the specific score, and * the scores used to compute the Total Score ($TS$).}
\label{tab:results}
    \begin{tabular}{|l|c|c|c|c|}
      \hline
      \textbf{Matching Score} & \textbf{TH} &  \textbf{P} & \textbf{R} &\textbf{F1} \\
      \hline
      Text Overlap ($TOS$) & 0.40 & 0.48 & 0.66 & 0.55 \\
      \hline
      Skill Overlap ($SOS$) * & 0.40 & 0.89 & 0.66 & 0.76 \\
      \hline
      Text Embedding ($TES$) & 0.50 & 0.52 & 0.81 & 0.63 \\
      \hline
      Skill Embedding ($SES$) * & 0.50 & 0.66 & 0.97 & 0.79 \\
      \hline
        Title Embedding ($TTES$)& 0.50 & 0.49 & 0.74 & 0.59 \\
      \hline
        All Text Embedding ($AES$) & 0.50 & 0.57 & 0.95 & 0.71 \\
      \hline
       Weighted Skill ($WSS$) * & 0.20 & 0.84 & 0.97 & 0.90 \\
      \hline
      \textbf{Total Score ($TS$)} & \textbf{0.35} & \textbf{0.89} & \textbf{0.99} & \textbf{0.94} \\
      \hline
    \end{tabular}
\end{table*}

\begin{figure*}[tbp]
  \centering
  \begin{subfigure}[b]{0.45\textwidth}
    \includegraphics[width=\textwidth]{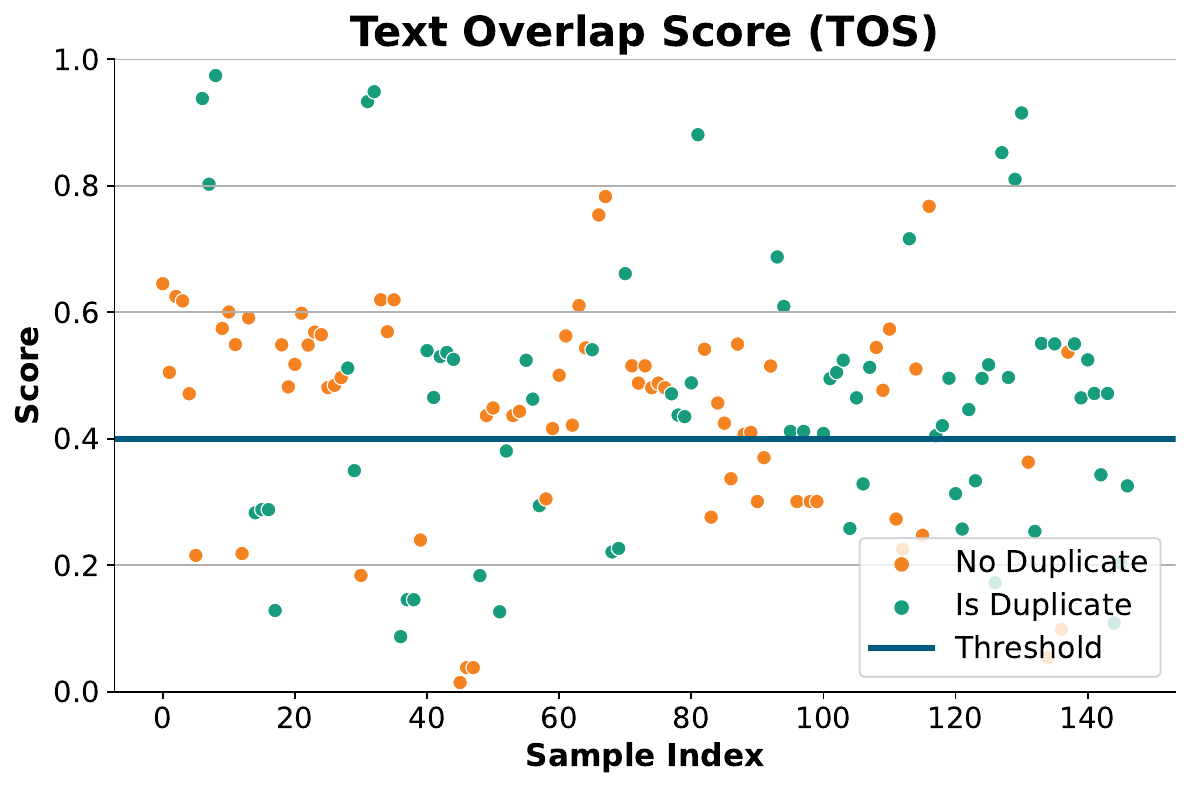}
    %\caption{Comparison of computed $\mathcal{L}_{WA}$ weights for base and fine-tuned models}
    \caption{}
    \label{fig:textscore}
  \end{subfigure}
  \hspace{4pt}
  \begin{subfigure}[b]{0.45\textwidth}
    \includegraphics[width=\textwidth]{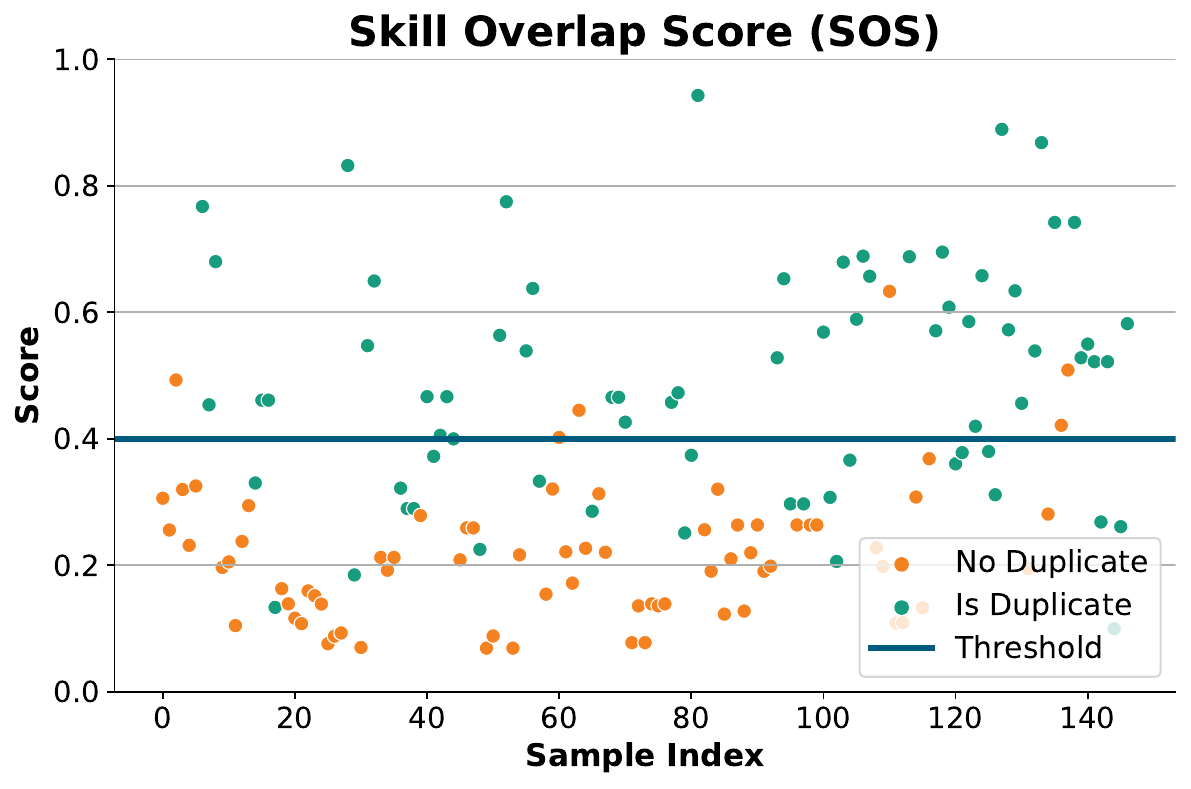}
    %\caption{Comparison of computed $\mathcal{L}_{WA}$ weights for leaflet and reports data set}
    \caption{}
    \label{fig:skillscore}
  \end{subfigure}
  \hspace{4pt}
  \begin{subfigure}[b]{0.45\textwidth}
    \includegraphics[width=\textwidth]{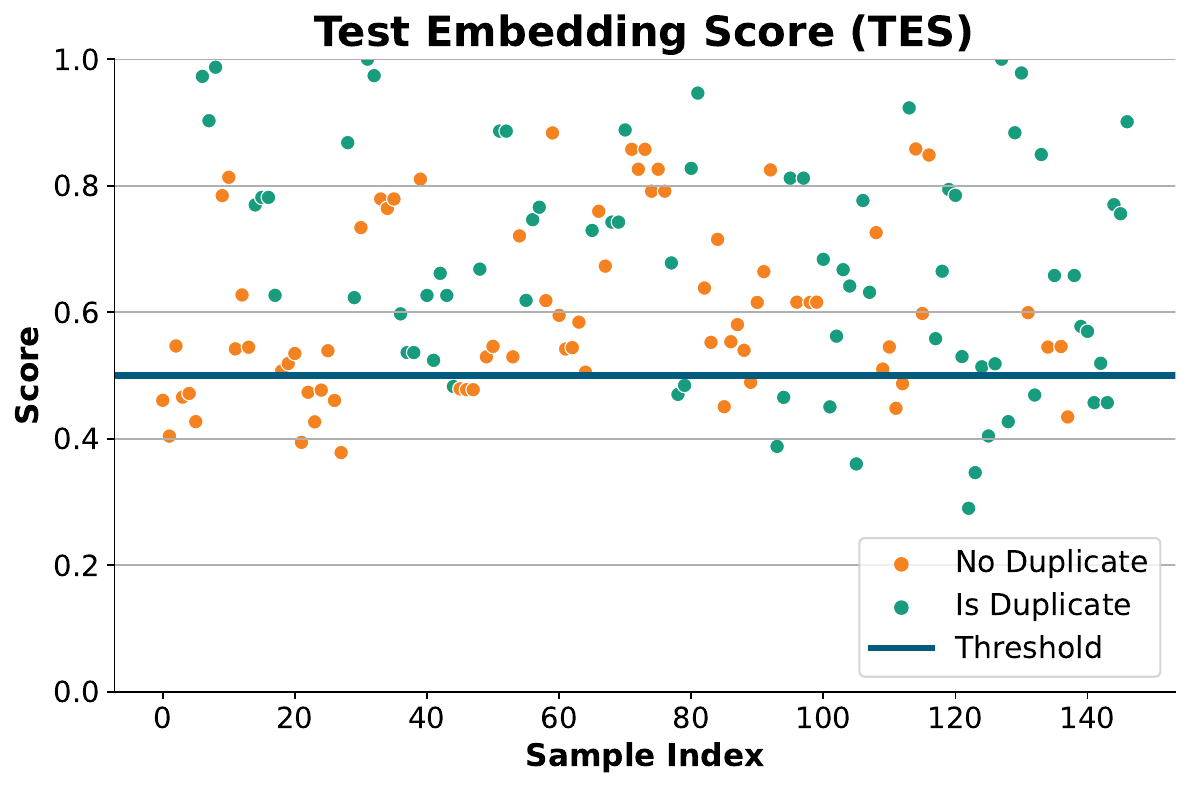}
    %\caption{Comparison of computed $\mathcal{L}_{WA}$ weights for leaflet and reports data set}
    \caption{}
    \label{fig:textembedding}
  \end{subfigure}
  \hspace{4pt}
  \begin{subfigure}[b]{0.45\textwidth}
    \includegraphics[width=\textwidth]{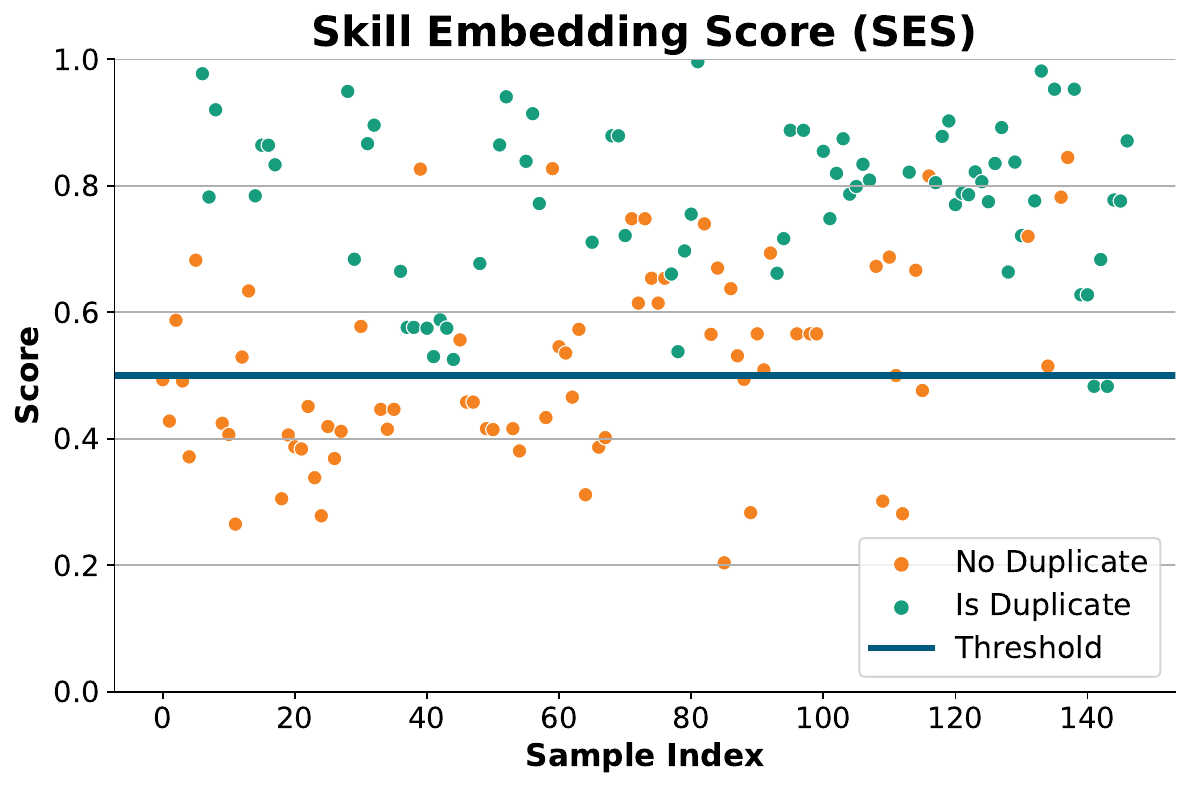}
    %\caption{Comparison of computed $\mathcal{L}_{WA}$ weights for leaflet and reports data set}
    \caption{}
    \label{fig:skillembedding}
  \end{subfigure}
  \hspace{4pt}
  \begin{subfigure}[b]{0.45\textwidth}
    \includegraphics[width=\textwidth]{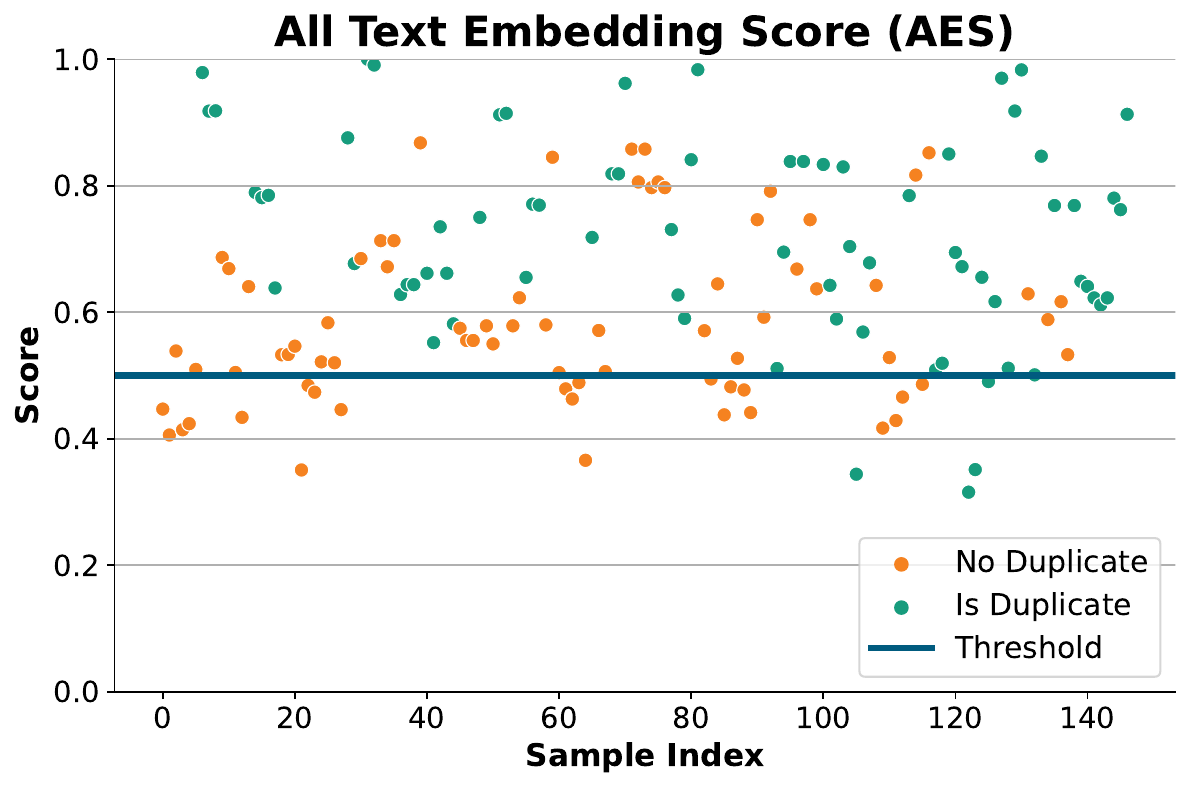}
    %\caption{Comparison of computed $\mathcal{L}_{WA}$ weights for leaflet and reports data set}
    \caption{}
    \label{fig:alltextembedding}
  \end{subfigure}
  \hspace{4pt}
  \begin{subfigure}[b]{0.45\textwidth}
    \includegraphics[width=\textwidth]{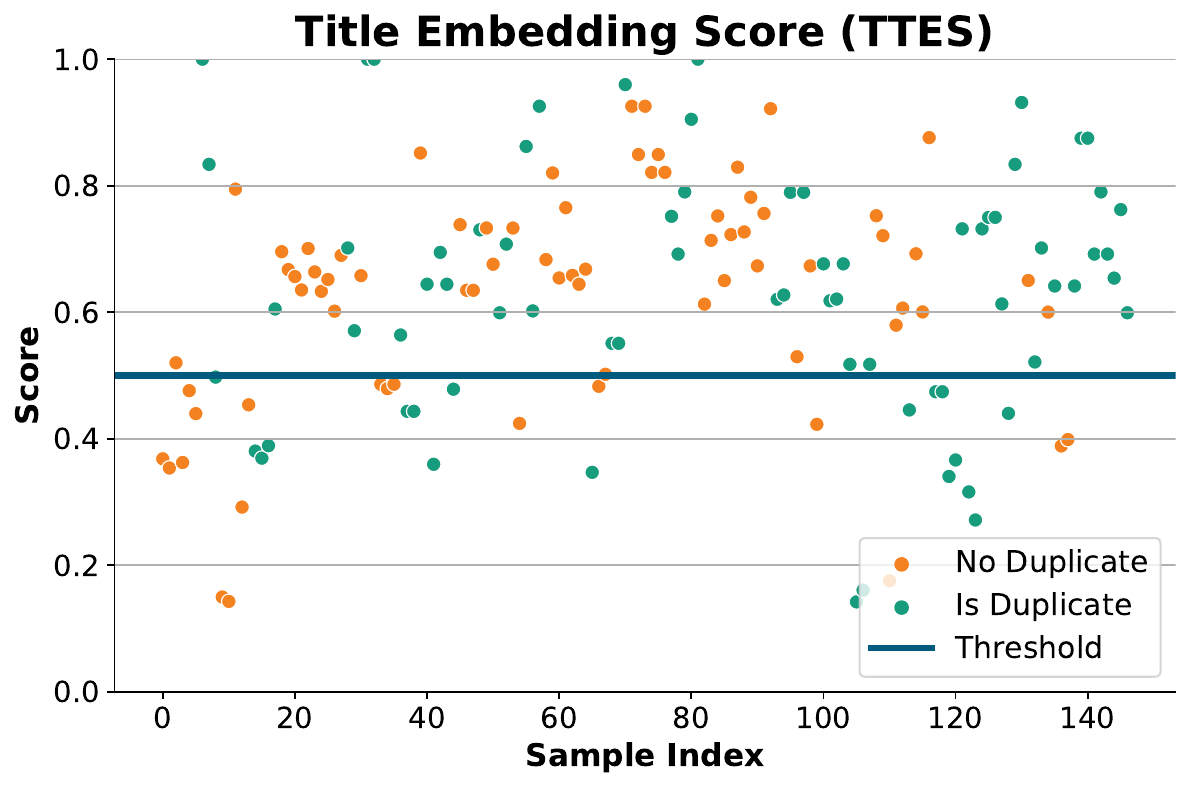}
    %\caption{Comparison of computed $\mathcal{L}_{WA}$ weights for leaflet and reports data set}
    \caption{}
    \label{fig:titleembedding}
  \end{subfigure}
  \hspace{4pt}
  \begin{subfigure}[b]{0.45\textwidth}
    \includegraphics[width=\textwidth]{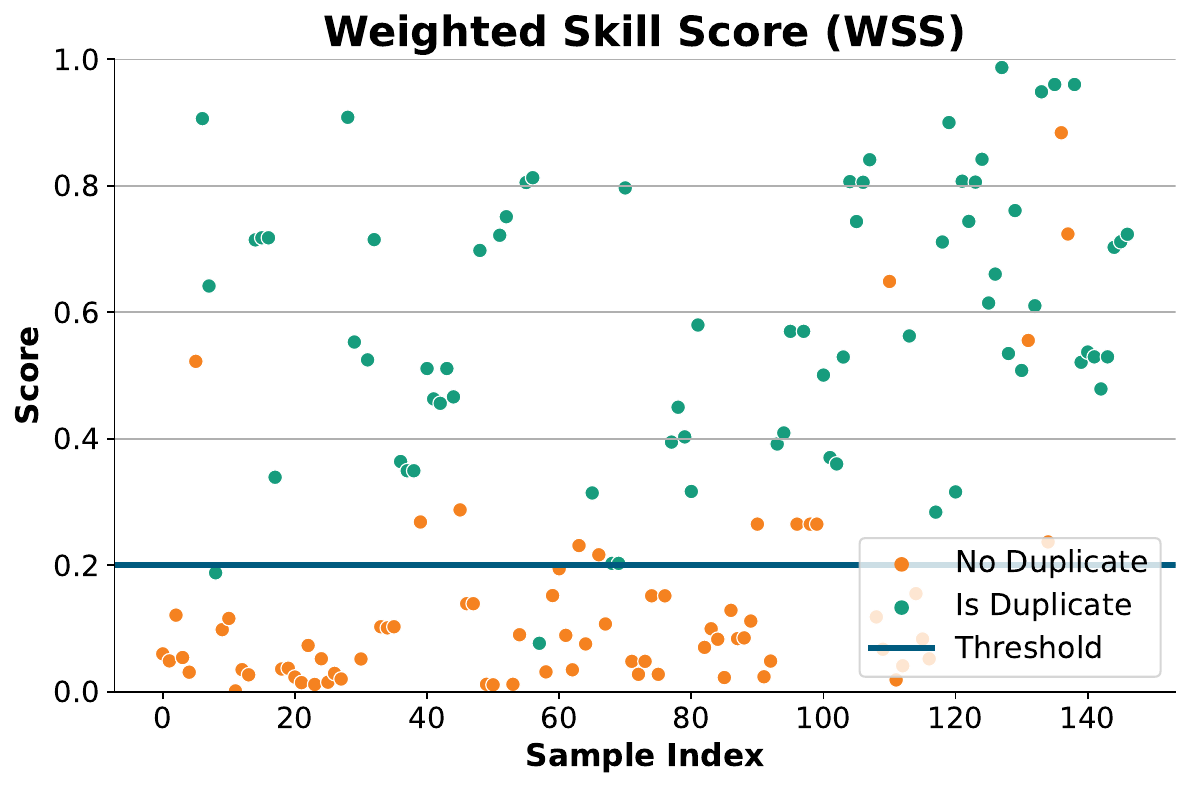}
    %\caption{Comparison of computed $\mathcal{L}_{WA}$ weights for leaflet and reports data set}
    \caption{}
    \label{fig:weightedskillscore}
  \end{subfigure}
  \hspace{4pt}
  \begin{subfigure}[b]{0.45\textwidth}
    \includegraphics[width=\textwidth]{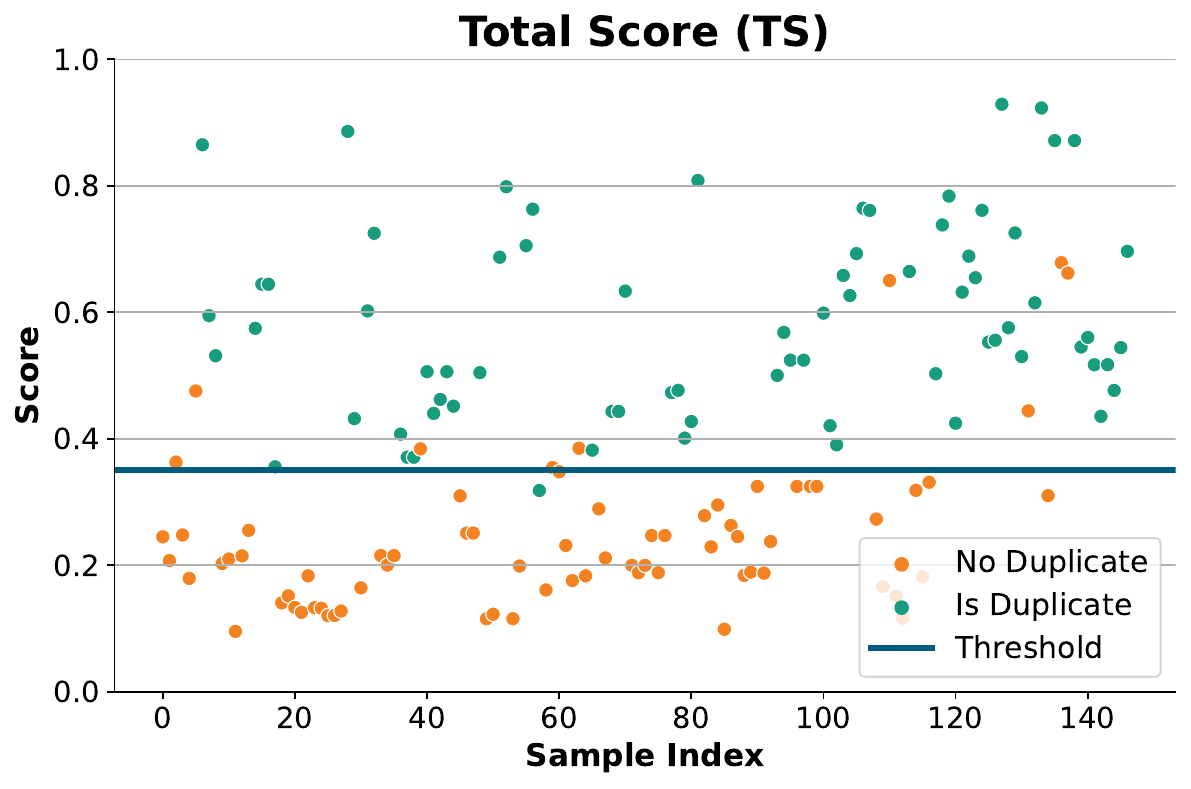}
    %\caption{Comparison of computed $\mathcal{L}_{WA}$ weights for leaflet and reports data set}
    \caption{}
    \label{fig:totalscore}
  \end{subfigure}
  \vspace{-1em}
  \caption{Score predictions for the different methods on the test set with 74 duplicates and 73 non-duplicates. The thresholds are set manually for each score to gain a good distinction between positive and negative duplicate predictions.}
  \label{fig:scores}
\end{figure*}

Comparing the performance of individual scores, Table~\ref{tab:results} and Figure~\ref{fig:scores} indicate that using the text overlap ($TOS$) and embedding score ($TES$) over the whole description texts is not sufficient for discriminating between real and fake duplicates. 
This is also due to the already mentioned existence of overlapping boilerplate text: some texts originate from the same company or even project, but indeed advertise different jobs.
%The latter ones might indeed have a lot of overlapping boilerplate text, which might be due to the fact that these texts originate from the same company or even project, but indeed are advertising different jobs.
When concentrating on the skill text containing only relevant technical terms specific for a given job offer by using the $SOS$ and $SES$ metrics, performance increases significantly.

In contrast, relying on the individual result of the title score $TTES$ is not helpful. Although in the beginning we expected the headline of a job offering to be a meaningful indicator for duplicate references, we realized that these texts without any further context are also highly misleading, since a lot of independent companies and projects use the same wording and technical terms (e.g. "SAP developer wanted") when publishing job postings.
Even combining the title and skill text with the original job description text in the all text embedding approach ($AES$) cannot deliver good performance. Overall, one of our key findings was that -- although they achieve the highest prediction scores for real duplicate pairs -- all of our embedding approaches in general seem not to be capable of separating fake duplicates from real ones, as the corresponding plots in Figure~\ref{fig:scores} underline.

On the other hand, only using the weighted skill score ($WSS$) without combining with any other measure already performs well. The fact that we attribute weights to significant skill features allows reliable detection of fake duplicates, since in contrast to the other metrics the influence of overlapping (but maybe misleading) boilerplate text on the prediction is reduced. 
Note that a requirement for the productive usage of this score computation is the continuous update of the underlying skill weights to capture the relevance of new terms and wordings that might arise over time. 

With combining the most promising individual scores of each group (namely the skill based scores) to form the total score $TS$, we can improve the performance on the validation set to an F1 score of $0.94$, meaning that most of the comparison pairs could be classified correctly into duplicates and non-duplicates. We believe that assembling  our scoring method in this way allows us to compensate  weaknesses of individual scores and to create more reliable and stable prediction results, as shown before \cite{kdir2023}.

During the final phase of the project, we applied this optimized setting to the test set consisting of $50$ samples which are exclusively real duplicate pairs. Since there are no non-duplicates in this dataset, the final test before productive deployment ensures that the constructed total score can identify the actual duplicates well by assigning them a relatively high score. 
Although there was a clearly visible scatter among predictions with a $TS$ ranging from $0.24$ to $0.93$, we could identify some indicators of similarity for every test sample, indicating the potential for further fine-tuning of the $TS$. For a threshold of $0.2$, all of the real duplicates would have been detected.
After feedback from the project partner, however, thresholds were not set to the optimal operating point for the (rather small) test set, but to a more robust value (see next section) for the final productive deployment. This threshold choice implies more importance to specificity (i.e. low false alarm rate) at the cost of sensitivity.
%Nonetheless, since manual inspection of results is time consuming, thresholds were not set to the optimal operating point for the (rather small) test set
%Nonetheless, since manual inspection of results is time consuming, thresholds were set differently in the final productive deployment, where only strong duplicate indications are considered, implying the risk of missing other duplicates with weaker signs.

\section{Productive Deployment}
After the experimental evaluation, the duplicate detection framework in deployed in production with periodic checks for recently published and collected job descriptions. In terms of scalability, once the initial vector database for the document embeddings is generated, the computational cost and time scales linearly in the number of processed documents. Since only a fixed time frame is compared to a new job posting, the number of comparisons can in fact vary but does not scale with the size of the dataset.

For the deployed system, the thresholds for the score computations were adapted and an additional criteria introduced: the $TS$ threshold is set to $0.6$ and each individual scoring method must achieve a minimum value of $0.1$ in order to classify a pair as duplicates.
%For live usage the thresholds for the score computations were adapted: Here we treat matching pairs having a minimum total score of $0.6$ and at least $0.1$ in every other individual scoring as detected duplicates.
With this setting, $58.860$ published job descriptions were analyzed and matched during a first productive phase of six months, of which $33.865$ were found to be no duplicates and $24.995$ to be duplicates of other projects. In detail, for every sample predicted as a duplicate we identified about $2.5$ duplicate job offerings on average.
This implies that the $24.995$ duplicate predictions contain about $10.000$ unique job postings, leading to a total sum of $43.705$ postings identified as unique within this period of time.
Overall, the positive feedback given by the end users of the tool after several months of productive usage confirm that our solution is working as intended and validates our approach.

\section{Limitations}
As stated by \citet{DraisbachN13}, choosing thresholds on a small test set of samples might not scale appropriately when the data set gets larger. Therefore, future experiments on larger (labeled) test sets are necessary for a more generalizable threshold estimation. This was however not possible in the scope of this work, due to the nature of requiring high-quality labeled data, which had to be collected manually by recruiting experts during the data collection phase of the project. 

An additional source of improvement is a more sophisticated determination of the weights assigned to the individual metrics that form the Total Score ($TS$). The assumption, that individual scoring approaches carry more information than others to solve a task in a specific domain, is reasonable and was also shown in \cite{kdir2023}.

\section{Conclusion and Future Work}
Our work showed that for duplicate detection in real world scenarios, like detecting clones of job descriptions, the integration of domain specific expert knowledge is crucial. While applying standard matching algorithms like string similarity and embedding comparison only leads to limited performance, we pointed out that combining these approaches with context specific additions -- in our case the emphasis of significant skill terms -- achieves satisfactory results (F1-score of 0.94) regarding capabilities for productive deployment.
For future work, we plan to do a larger evaluation on bigger (human-curated) test sets to identify possible weak points of our approach that could not be addressed within the limited scope of this first project. \\
Furthermore, as large language models (LLMs) become more advanced even in not explicitly shown domains, they can be used as a baseline classifier in zero- or many-shot evaluations. This allows for a more comprehensive evaluation, with the LLM serving as a proxy for a human evaluator. Additionally, available LLMs can be fine-tuned on the recruiting domain and used as an additional scoring method, either in parallel or incorporated as on component in the Total Score ($TS$). However, this raises the challenge of keeping the tool computationally efficient, as SOTA LLMs are quite expensive to operate. \cite{EnergyCosts_Samsi2023} \\
Incorporating feedback mechanisms for users is currently done in an offline fashion by utilizing their feedback on classification performance to re-calibrate the thresholds over time. This could also be done automatically at predefined intervals or by using distribution shift \cite{distShiftDetection_NEURIPS2019} or performance monitoring approaches \cite{Ginart2021MLDemonDM}.

\backmatter

%%\bmhead{Supplementary information}

%%\bmhead{Acknowledgments}

\bmhead{Disclosure of Interests}
The authors have no competing interests to declare that are relevant to the content of this article.

\bibliography{sn-bibliography}% common bib file
%% if required, the content of .bbl file can be included here once bbl is generated
%%\input sn-article.bbl

\end{document}